\documentclass[sigconf]{acmart}

\usepackage{algorithm}
\usepackage{algorithmic}
\usepackage{multirow}
\usepackage{subfigure}
\AtBeginDocument{%
  }

\acmConference[MM '23]{Proceeding of the 31th ACM International Conference on Multimedia}{October 29-- November 2, 2023}{Ottawa, Canada}

\renewcommand\footnotetextcopyrightpermission[1]{}
\begin{document}

\title{All-in-one Multi-degradation Image Restoration Network via Hierarchical Degradation Representation}

\author{Cheng Zhang}
\email{zhangcheng233@mail.nwpu.edu.cn}
\affiliation{%
  \institution{Northwestern Polytechnical University}
  \city{Xi'an}
  \country{China}
}

\author{Yu Zhu}
\email{yuzhu@nwpu.edu.cn}
\affiliation{%
  \institution{Northwestern Polytechnical University}
  \city{Xi'an}
  \country{China}
}
\authornote{Corresponding author}

\author{Qingsen Yan}
\email{qingsenyan@nwpu.edu.cn}
\affiliation{%
  \institution{Northwestern Polytechnical University}
  \city{Xi'an}
  \country{China}
}
\author{Jinqiu Sun}
\email{sunjinqiu@nwpu.edu.cn}
\affiliation{%
  \institution{Northwestern Polytechnical University}
  \city{Xi'an}
  \country{China}
}
\author{Yanning Zhang}
\email{ynzhang@nwpu.edu.cn}
\affiliation{%
  \institution{Northwestern Polytechnical University}
  \city{Xi'an}
  \country{China}
}

\begin{abstract}
The aim of image restoration is to recover high-quality images from distorted ones. However, current methods usually focus on a single task (\emph{e.g.}, denoising, deblurring or super-resolution) which cannot address the needs of real-world multi-task processing, especially on mobile devices. Thus, developing an all-in-one method that can restore images from various unknown distortions is a significant challenge. Previous works have employed contrastive learning to learn the degradation representation from observed images, but this often leads to representation drift caused by deficient positive and negative pairs. To address this issue, we propose a novel All-in-one Multi-degradation Image Restoration Network (AMIRNet) that can effectively capture and utilize accurate degradation representation for image restoration. AMIRNet learns a degradation representation for unknown degraded images by progressively constructing a tree structure through clustering, without any prior knowledge of degradation information. This tree-structured representation explicitly reflects the consistency and discrepancy of various distortions,
providing a specific clue for image restoration.
To further enhance the performance of the image restoration network and overcome domain gaps caused by unknown distortions, we design a feature transform block (FTB) that aligns domains and refines features with the guidance of the degradation representation.
We conduct extensive experiments on multiple distorted datasets, demonstrating the effectiveness of our method and its advantages over state-of-the-art restoration methods both qualitatively and quantitatively.
    \vspace{-0.1cm}
\end{abstract}




\maketitle
\newcommand{\eg}{\textit{e}.\textit{g}. }
\section{Introduction}
Image restoration is a critical topic in low-level vision, which generates a high-quality image from a damaged image caused by degradation, \eg , blurriness, noise, and low illumination.
While a particular type of degradation typically prevails in an image, it is common to encounter situations where multiple degradations \footnote{Following \cite{li2022all}, the multi-degradation that this paper focuses on refers to a dataset with multiple degradations, differing from mixed degradations in an image.} need to be processed in the real world.
For instance, such a scenario may arise when capturing photography with different camera parameters \cite{chen2018learning,zhang2022exploring},  collecting image data from the internet \cite{ma2016waterloo}, and  taking pictures under adverse weather conditions \cite{valanarasu2022transweather}.

The majority of the existing restoration methods \cite{xu2017multi,abuolaim2020defocus,zamir2021multi,liang2021swinir,chen2022simple,chen2021hinet,guo2019toward,zhang2017beyond, zhang2020attention, yan2023sharpformer} are designed for a single degradation (\emph{e.g.} denoising, deblurring, or super-resolution), and even some methods \cite{zamir2021multi,liang2021swinir,chen2022simple} that claim to handle multiple degradations require separate training on categorized and specific degraded images, which is computationally expensive, time-consuming for optimization, and not friendly for storage on mobile devices. Moreover, specifying the degradation types adds complexity to the usage and increases the risk of performance degradation. Therefore, considering these issues, an all-in-one approach is the optimal solution for restoring images with multiple unknown degradations, as it allows for unified training and convenient testing with the same parameters and architecture, thus reducing complexity and increasing ease of use.

Within the all-in-one framework, the crucial problem to be tackled is how to represent and leverage the degradation information in the restoration network, since improved representation leads to enhanced restoration performance, particularly in situations involving multiple degradations.
Some degradation estimation methods \cite{liu2013single,guo2019toward,gong2017motion,pan2016robust,chen2021blind,hu2014deblurring,yan2017image} typically assume a predefined degradation category and estimate degradation level parameters, making them less suitable for scenarios with multiple unknown degradations. While some methods \cite{wang2021unsupervised, li2022all} rely on contrastive learning for degradation representation, the deficient selection of positive and negative samples in these approaches makes it difficult to comprehensively describe the relationships between different degradations, which can lead to representation drift and ultimately impacts the performance of image restoration.

\begin{figure}[h]
  \centering
  \includegraphics[width=0.85\linewidth,trim=1 1 1 1,clip]{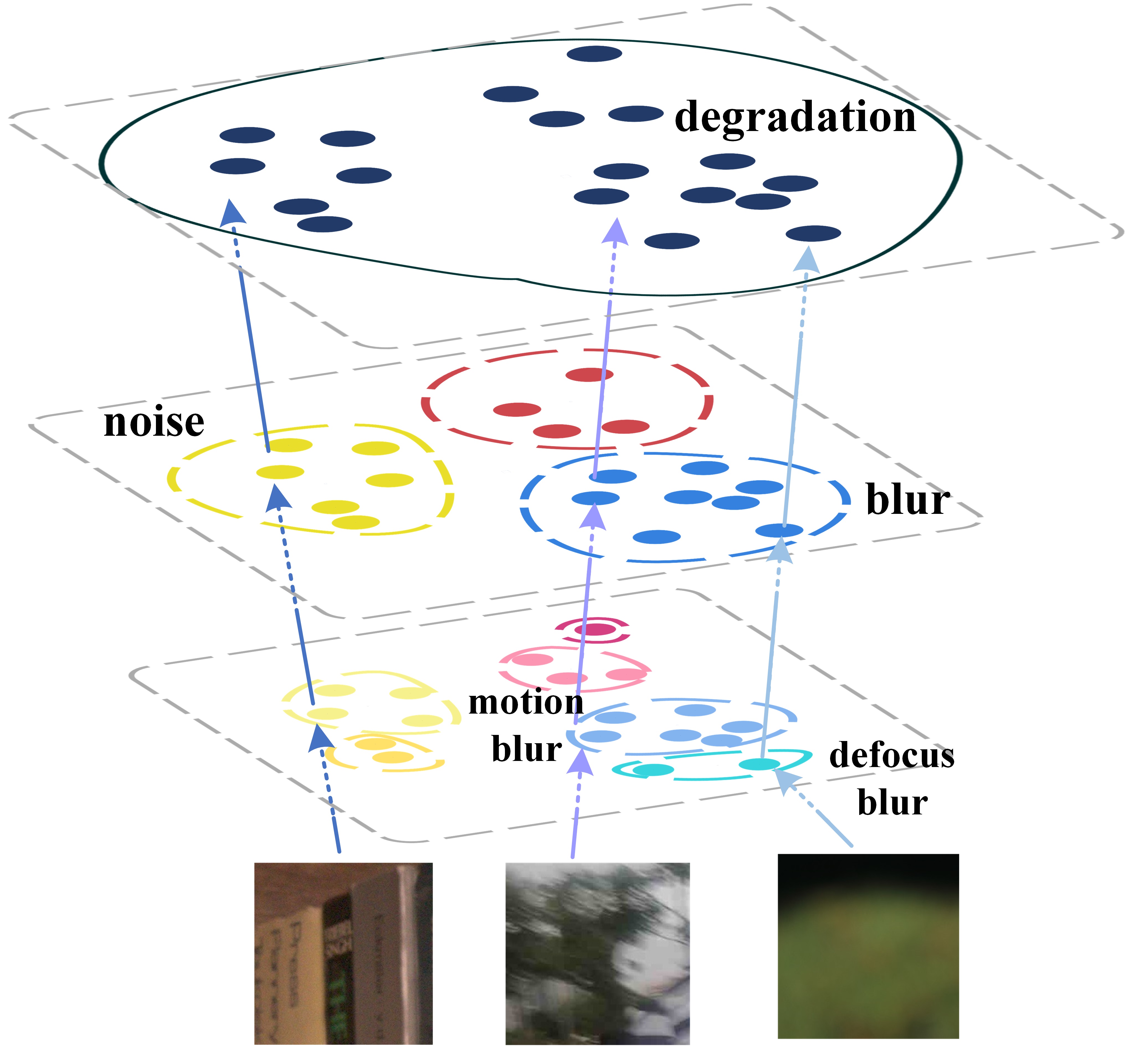}
  \caption{Degradation possesses a characteristic of hierarchical subordination \emph{e.g.} degradation $\rightarrow$ blur $\rightarrow$ motion blur from coarse to fine. As presented in the figure, two blurred images may be clustered as blur, but at a lower level of the hierarchy, they are distinguished as motion blur and defocus blur.}
  \label{fig:teaser}
  \vspace{-0.2cm}
\end{figure}

To tackle the aforementioned challenges, we propose a novel All-in-one network, named AMIRNet, to handle multi-degraded images via learning hierarchical degradation representation. Specifically, we observe that the degradations exhibit a characteristic of hierarchical subordination, as illustrated in Figure \ref{fig:teaser}. As an example, in a dataset with multiple degradations, two images may be grouped together based on the presence of blur, but a more specific lower-level cluster can further categorize them into defocus blur and motion blur, respectively. 
The hierarchical structure enables the modeling of commonalities and distinctions among image degradations, which provides a beneficial clue for all-in-one image restoration under multiple degradations. 

Therefore, we propose a tree-structured representation to capture the relationship between multi-degradations and progressively construct the representation through clustering from coarse to fine. Similar to training networks separately on individual degraded datasets \cite{zamir2021multi,liang2021swinir,chen2022simple}, the use of the hierarchical degradation representation is aimed at making feature distributions of similar degradations more compact and easily distinguishable from dissimilar degradation features.
Additionally, to sufficiently leverage the degradation representation in the all-in-one restoration network, we devise a feature transform block (FTB) to integrate the image feature and corresponding degradation representation. 
Considering the domain gap caused by distortions, we draw inspiration from \cite{chang2019domain} and incorporate degradation-related layer normalization in the FTB to align domains. We also introduce a degradation-related gating mechanism in the FTB to control the information flow in the restoration network. The FTB can refine image features, enabling the network to adapt to various degraded images. Furthermore, we conduct comprehensive experiments and ablation studies on multi-degradation datasets to verify the effectiveness of our method.

Our main findings and contributions can be summarized as follows:
\begin{itemize}
\vspace{-0.1cm}

\item { We propose a novel All-in-one Multi-degradation Image Restoration Network (AMIRNet) to handle adverse degraded images in the real world.}
\item {Based on the observation that multi-degraded images follow a hierarchical structure, we propose to progressively construct a tree-structured representation by clustering to characterize the similarity and difference between degradations.}
\item {We devise a feature transform block (FTB) to overcome domain gaps caused by various distortions and refine image features with the guidance of degradation representation.}

\item{Extensive experiments are conducted to confirm the effectiveness of our method, which not only has the potential for promising results but also achieves state-of-the-art performance on multi-degradation datasets.}
\end{itemize}

\begin{figure*}[htbp]
  \centering
  \includegraphics[width=\linewidth,trim=3 3 3 3,clip]{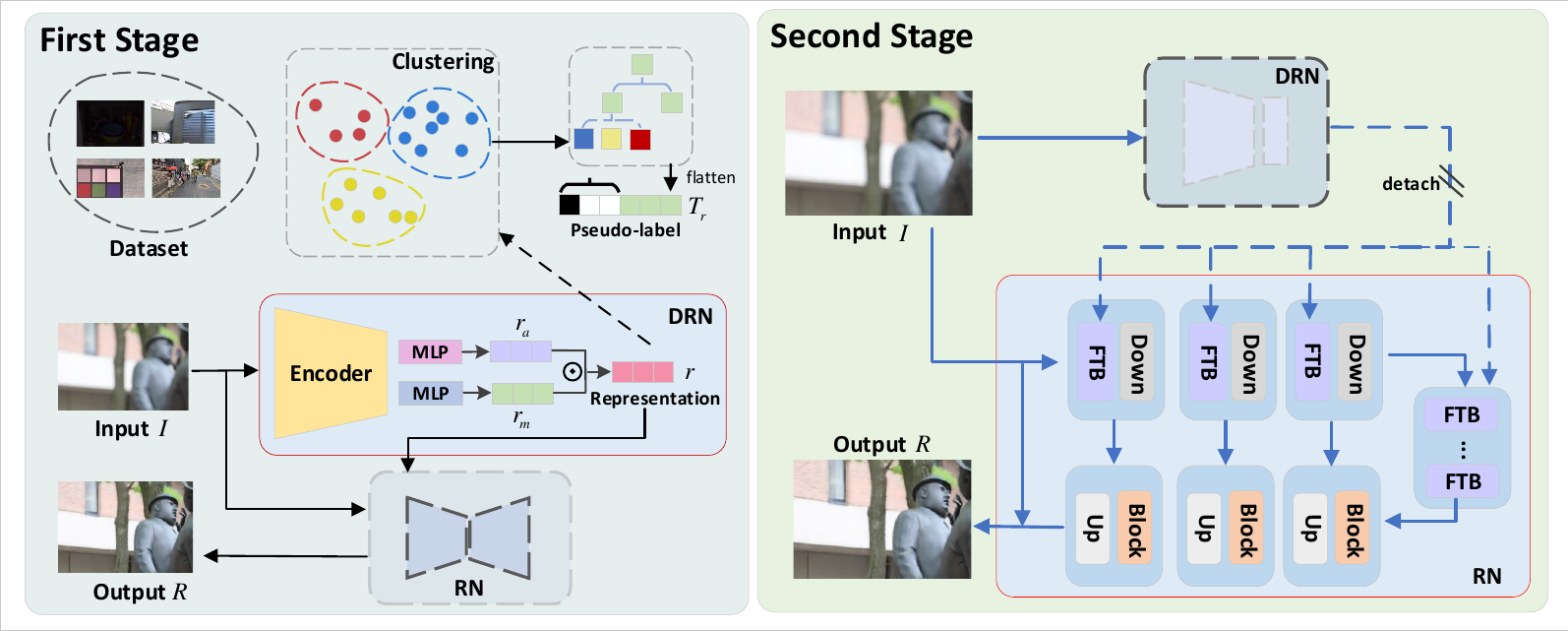}
  \caption{The overview of our proposed All-in-one Multi-degradation Image Restoration Network (AMIRNet), which consists of two main components, the degradation representation network (DRN) for learning hierarchical degradation representation and the restoration sub-network (RN) for recovering a high-quality image from a distorted image. In the first stage, we focus on constructing the degradation representation and training the DRN. In the second stage, we fix the parameters of the DRN and retrain the RN.}
  \label{fig:overview}
  \vspace{-0.3cm}
\end{figure*}

\vspace{-0.3cm}
\section{Related works}

\subsection{Single Degradation Restoration}
Image restoration is a fundamental task in computer vision, which aims to recover the degraded images to their original high-quality versions, including deblurring, denoising, inpainting, low-light enhancement, and so on.
Traditional approaches focus on the exploration of the image prior, such as sparse \cite{luo2015removing,mairal2007sparse,xu2013unnatural}, low-rank \cite{gu2014weighted,xu2017multi}, self-similarity \cite{dabov2007image} \textit{etc.}. 

Recently, with the support of a large number of collected paired images, many deep neural networks (DNN) methods \cite{abuolaim2020defocus,liang2021swinir,zamir2021multi,chen2022simple, Chen2018Retinex, chen2021hinet, zhang2020attention, yan2023sharpformer, yan2019attention} have produced impressive results on each subtask of restoration. Those works emphasize the design of network architecture and loss functions. By utilizing images that are usually captured and categorized manually by humans according to degradation type, DNN-based methods focus on learning the implicit mapping between the distorted image and the high-quality image.
Although some models \cite{zamir2021multi,liang2021swinir,chen2022simple,chen2021hinet} can be adapted to handle multiple types of degraded images, they typically require separate training on specific degraded datasets, and they may not generalize well to other typed of degraded images without further adaptation.
For instance, MPR \cite{zamir2021multi} trained well for image denoising has limited performance on image deblurring, which is not expected in practice. Therefore, it is vital to consider the fact that images are often corrupted by multiple degradations and devise all-in-one solutions to meet the requirement of real-world multi-task processing.

\subsection{Multi-degradation Restoration}
Currently, there has been increased interest in developing all-in-one models that can handle various degraded images in a single network after being trained. For instance, \cite{li2020all} proposes a network with multiple encoders to process each degradation using a specific encoder. Similarly, Transweather \cite{valanarasu2022transweather} introduces a decoder with learnable embeddings in a transformer architecture to address multiple degradation types.

Instead of only focusing on designing network structures, there are also techniques that incorporate contrastive learning to enable the network to differentiate between different types of corruptions and handle multi-degraded images. One such example is presented in \cite{chen2022learning}, which suggests using both soft and hard contrastive regularization to enhance the performance of both specific and multiple degradations. Other approaches, DASR \cite{wang2021unsupervised} and AirNet \cite{li2022all} consider patches from the same image as positive samples and patches from different images as negative samples. However, this selection of positive and negative samples may not always be adequate, since it does not take into account situations where different degradations are related or where different images may belong to the same degradation, which could result in representation drift and performance drop of restoration network. 

\subsection{Degradation Representation} 
The representation of degradation is a crucial step in image restoration and serves as a prerequisite for accurately restoring a degraded image. Generally, the type of degradation is known before the estimation and the main task is to estimate the parameters of the degradation model. For example, in denoising \cite{liu2013single,guo2019toward}, the core is to estimate the noise level when the noise type is known.
And in deblurring, many methods \cite{gong2017motion,pan2016robust,chen2021blind,hu2014deblurring,yan2017image} usually estimate the blur kernel before non-blind deblurring. 
Additionally, there exist approaches that implicitly represent the degradation by learning a feature vector that serves as a proxy for the degradation and is subsequently fed into the restoration method. DASR \cite{wang2021unsupervised} and AirNet \cite{li2022all} realize the learning of representation by contrastive learning. \cite{li2022learning} proposes to learn degradation representations with a blurry-sharp cycle framework. \cite{li2022learningsuper} proposes to learn a latent representation space for degradations in super-resolution. Our method falls into this category, in which we guide the network to restore clear images by constructing hierarchical degradation representations of degraded images and adapting to different degraded image features.

\section{Methodology}

We aim to develop an \textbf{All-in-One} model that can effectively handle multiple types of degraded images, eliminating the requirement for retraining or fine-tuning once the training process is complete. As mentioned above, the primary obstacle in developing an all-in-one method for multi-degradation is how to represent and leverage the degradation information within the restoration network. In this section, we will address the issue and present a detailed explanation of our proposed solution.

\subsection{Overview}
Our solution is a two-stage All-in-One approach, the first stage is designed to construct the hierarchical representation of degraded images, and the second stage is to remove degradation artifacts to produce a high-quality image with the supervision of degradation representation. The overview of our method is depicted in Figure \ref{fig:overview}. Despite serving different purposes, the two stages share a common network architecture. The network we propose comprises of two main components: a Degradation representation sub-Network (DRN) and a restoration sub-network (RN). Initially, a corrupted image $I$ is fed into the encoder of the degradation representation sub-network to extract its features $z$, which are then transformed by projectors to yield the hierarchical degradation representation $r$ in latent space. Subsequently, the degradation representation $r$ is integrated into the restoration via the newly proposed feature transform block. With the assistance of the degradation representation $r$, our network is able to recover a high-quality image from the input distorted image.

\begin{algorithm}[htbp]
\caption{Degradation Representation Construction }
\label{algo:spc}
\begin{algorithmic}[1]
\REQUIRE Image pairs $D = {(I_1,Y_1), (I_2,Y_2), ..., (I_n,Y_n)}$,
    number of hierarchical layers $L$,
    number of clusters in each layer $C = {c_1, c_2, ..., c_L}$
\ENSURE Tree-structured representation $T = {T_1,T_2,...,T_n}$
\STATE Initialize $T$ with the root node
\FOR{$i=1$ to $L$}
\FOR{$j=1$ to $c_i$}
\STATE Select $CurrentNode$  $N_{i,j}$
\STATE  Obtain $r_k$, $k \in N_{i,j}$
\STATE Cluster $T_{k,i} \gets KMeans(r_k)$
\STATE Update $T_k \gets concat(T_k,T_{k,i})$
\ENDFOR
\STATE Optimize DRN and RN 
\ENDFOR
\STATE \textbf{return} Tree-structured representation $T = {T_1,T_2,...,T_n}$
\end{algorithmic}
\end{algorithm}
\vspace{-0.4cm}

\subsection{Hierarchical Degradation Representation}
As mentioned above and shown in Figure \ref{fig:teaser}, the degradation has a characteristic of hierarchical subordination. The characteristic indicates that the images with adverse distortions can be classified into different categories in each layer of the hierarchy. 
An instance could be a motion-blurred image and a defocused image both categorized under "blur", however, their classification in the subgroup of blur types is different.
To model the commonality and the distinction between multiple distortions, naturally, we propose a novel tree-structured representation, which corresponds to the categories of the hierarchical structure of degradations.
The tree-structured representation $T$ can be flattened into a vector when performing a level-order traversal of the hierarchy, which is formulated as follows:
\begin{equation}
    T = [v_{1,1},..., v_{i,j},...,v_{L,c_L}], v_{i,j} \in \{0,1\} ,
\end{equation}
where $L$ and $c_L$ denote the number of levels in the tree structure and the number of nodes in each level, respectively.
The binary value $v_{i,j}$ in $j_{th}$ node of $i_{th}$ level represents whether the image belongs to that node or not. In our method, we predefine the structure of the tree as a 4-layer binary tree, so that it can be used consistently during both training and testing. After establishing the specific form of the hierarchical representation, we need to take into account how to construct the representation for each corrupted image without additional degradation information and learn it with a degradation representation network.

 \textbf{Representation Construction.} We recommend a progressive strategy to construct the degradation representation from the top layer to the bottom layer in the hierarchy, as described in Algorithm \ref{algo:spc}. Given the corrupted images $I$, clear images $Y$, the number of hierarchy $L$, and the number of clusters in each layer $c_1,c_2,...,c_L$, our goal is to build a tree-structured representation $T$.
 The degradation representation of each input image is initially assigned to the root node and updated with the increasing depth of the tree during the outer loop. In each iteration of a layer, a node is sequentially selected as the current node, and the samples belonging to the current node are located to obtain their degradation representation using DRN.
 After clustering, the clustering results are concatenated with the original representations to update the degradation representations.

\textbf{Representation Learning.} To learn the degradation representation and facilitate usage during testing, a sub-network DRN is employed to extract degradation features and transform them into a low-dimensional embedding space under the supervision of the hierarchical tree-structured representation. Our DRN includes an encoder and two parallel projectors. The encoder $\mathcal{F}$ is composed of multiple convolutional layers with residual connections, activation layers, and pooling layers. The encoder's objective is to extract features about degradations from degraded image $I$, and the process is expressed as $z=\mathcal{F}(I)$.
The projectors are based on MLPs, which are composed of fully-connected layers and activation functions.
One of the projectors, named mask projector $\mathcal{P}_m$, is designed to predict the binary mask $r_m=\mathcal{P}_m(z)$.
The length of vector $r_m$ is always consistent with the current clustering layer in the construction process of tree-structured representation and will increase with the number of nodes in the degradation tree.

Given a degraded image $I$ and its degradation $r$, we calculate Cross Entropy loss to optimize the parameters of the encoder $\mathcal{F}$ and the mask projector $\mathcal{P}_m$:
\begin{equation}
    \mathcal{L}_{cls} = CrossEntropy(r_m, T).
\end{equation}

And the other parallel projector, named attribute projector $P_a$ is only used in the first training stage, its output has the same size as the output of the first projector. The value on each vector dimension represents the attribute value at the corresponding node. In the first training stage, the degradation representation can be described as a product of mask $r_m$ and attribute value $r_a=\mathcal{P}_a(z)$ :
\begin{equation}
    r = r_m \cdot r_a .
\end{equation}

\begin{table*}[htbp]

\caption{Quantitative comparison of our method and other state-of-the-art methods on multi-degraded image datasets.
The datasets include various degradations such as blur, noise, low illumination, and defocus. All results are measured in terms of PSNR, SSIM, and LPIPS \cite{zhang2018unreasonable}. The red color indicates the best results, and the blue color indicates the second-best results.}
\setlength\tabcolsep{2pt}
\begin{tabular}{c|c|c|lll|lll|lll|lll|lll}
\hline
\multirow{2}{*}{Model} & \multirow{2}{*}{Param} & \multirow{2}{*}{FLOPs}          & \multicolumn{3}{c|}{Average}                   & \multicolumn{3}{c}{RED4\cite{Nah_2019_CVPR_Workshops_REDS}} &\multicolumn{3}{c}{SIDD\cite{abdelhamed2018high}} &\multicolumn{3}{c}{LOL\cite{Chen2018Retinex}} &\multicolumn{3}{c}{DPDD\cite{abuolaim2020defocus}}                         \\ \cline{4-18} 
                              & & & \multicolumn{1}{c}{PSNR} & \multicolumn{1}{c}{SSIM} & \multicolumn{1}{c|}{LPIPS} &
                                 \multicolumn{1}{c}{PSNR} & \multicolumn{1}{c}{SSIM} & \multicolumn{1}{c|}{LPIPS} &
                                 \multicolumn{1}{c}{PSNR} & \multicolumn{1}{c}{SSIM} & \multicolumn{1}{c|}{LPIPS} &
                                 \multicolumn{1}{c}{PSNR} & \multicolumn{1}{c}{SSIM} & \multicolumn{1}{c|}{LPIPS} &
                                \multicolumn{1}{c}{PSNR} & \multicolumn{1}{c}{SSIM} & \multicolumn{1}{c}{LPIPS} \\ \hline
          SwinIR  & 11.46 &  752.1          & 27.97                   & 0.7797                  & 0.3725 &     25.58               &  0.7297 &   0.4084         & 36.24 & 0.9019 & 0.3149 & 17.74 & 0.7509 &0.3989 & 24.59& 0.7267 &0.3945  \\
    MPR      &20.13 & 1707.4               & \textcolor{blue}{28.89}                   & \textcolor{blue}{0.8098}     &0.3540             & 26.53                   & 0.7599        &   0.4070 & 37.54 & 0.9125  &0.3267 & \textcolor{blue}{22.72} &\textcolor{red}{0.8406} & \textcolor{red}{0.2537} & \textcolor{blue}{25.10} &\textcolor{blue}{0.7641}      & 0.3628\\
    NAFNet        & 17.06 &  15.97            & 28.87                   & 0.8089 &\textcolor{blue}{0.3302}                   & \textcolor{red}{26.80} & \textcolor{blue}{0.7611} & \textcolor{blue}{0.3969}&\textcolor{blue}{37.80}                   & \textcolor{blue}{0.9211} &\textcolor{blue}{0.2614}             & 22.29 & \textcolor{blue}{0.8280} &\textcolor{blue}{0.2658} & 24.91 &0.7584 &\textcolor{blue}{0.3565}    \\
    TransWeather     & 37.68 &   6.13    & 26.90                   & 0.7508   &0.4062                & 25.03                   & 0.7210       &0.4358   & 34.59 & 0.8571 &0.4231& 21.58 & 0.8071 &0.2748& 23.49 & 0.6979 &0.3979         \\
    AirNet         & 5.77 &    301.3       & 28.29                   & 0.7840     &     0.3796          & 25.82                   & 0.7360  &0.4125                & 37.25  & 0.9030 &0.3041 & 13.79  & 0.7166 & 0.3639& 24.78 & 0.7342 & 0.4136\\
Ours          & 71.76 &   73.23           & \textcolor{red}{29.39}          & \textcolor{red}{0.8204}     & \textcolor{red} {0.3204}      & \textcolor{blue}{26.77}         & \textcolor{red}{0.7671}   &   \textcolor{red}{0.3875}   &\textcolor{red}{38.46}          & \textcolor{red}{0.9280}& 
\textcolor{red}{0.2608} &\textcolor{red}{22.83}          & 0.8259 & 
0.2662 & \textcolor{red}{25.45}          & \textcolor{red}{0.7739}  & \textcolor{red}{0.3412}\\ \hline
\end{tabular}
\label{tab: sota-results}
\end{table*}

\begin{figure}[htbp]
  \centering
  \includegraphics[width=\linewidth,trim=1 1 1 1,clip]{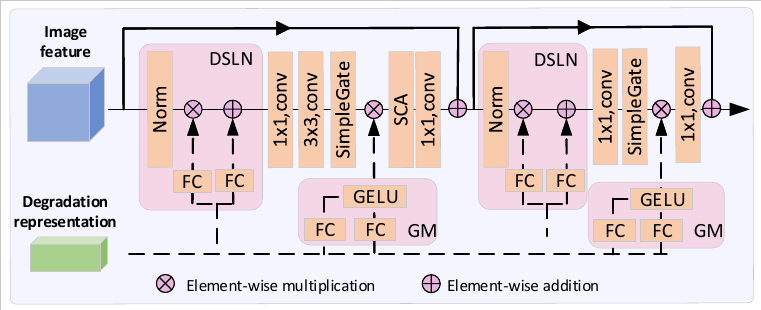}
  \caption{The architecture of feature transform block (FTB) used in the restoration network. FTB is employed to align the domain gaps caused by adverse distortions and refine degraded image features.}
  \label{fig:block}
\end{figure}

\subsection{Restoration with Degradation Guidance}
\textbf{Feature Transform Block.} To leverage the hierarchical degradation representation in image restoration, we introduce a feature transform block to modulate the image feature transformation in latent feature space according to the distortion information. The structure of this block is depicted in Figure \ref{fig:block}.
Inspired by Domain-Specific Batch Normalization (DSBN) \cite{chang2019domain} in domain adaptation, which transforms domain-specific information into domain-invariant representation using the parameters of BN, we employ degradation-specific parameters in layer normalization (LN) to refine the degraded image features. LN is a widely used technique in restoration networks, which is expressed as
\begin{equation}
    \text{LN}(x) = \frac{x - \mu}{\sqrt{\sigma^2 + \epsilon}} * \gamma + \beta ,
\end{equation}
where $\mu$ and $\sigma$ denote the mean and standard deviation of the image feature $x\in \mathbb{R}^{C\times H\times W}$, and $\epsilon$ is a small constant to avoid dividing by zero, and $\gamma$ and $\beta$ are learnable affine parameters.
In order to make the layer normalization adapt to various degraded image features, we propose a Degradation-Specific Layer Normalization (DSLN). Formally, DSLN allocates degradation-specific affine parameters $\gamma(r)$ and $\beta(r)$ for different degradations.

Hence, DSLN can be written as:
\begin{equation}
    \text{DSLN}(x) = \frac{x - \mu}{\sqrt{\sigma^2 + \epsilon}} * \gamma(r) + \beta(r) ,
\end{equation}
where $\gamma(r)$ and $\beta(r)$ are connected with the degradation representation $r$ by using linear transformation matrixes, which is realized by a fully-connection layer in the implementation,i.e., $\gamma(r) = Wr + b$.

Moreover, we incorporate a Gating Mechanism (GM) to activate the channels of image features according to degradation. The gating mechanism is formulated as the element-wise product of image feature $x$ and $\phi(r)$, and $\phi(r) = GELU(W^1r+b^1)(W^2r+b^2)$. The gating mechanism controls the information flow with the guidance of degradation representation $r$, thereby allowing the network to focus on the degradation-specific channels. 
Overall, with DSLN and gating mechanism, the FTB has the capability to align degradation domains and refine the image features under the guidance of degradation representation $r$, allowing the restoration network to produce high-quality images.

\textbf{Training and Loss.} Our restoration network is constructed based on NAFNet \cite{chen2022simple}, which is a variant Unet with symmetric encoder-decoder architecture. In the encoder of our restoration network, the FTB serves as the fundamental module and modifies the degradation-related image features. In the decoder, efficient NAFBlocks \cite{chen2022simple} are employed to transform modulated image features. The encoder features are concatenated with the decoder features via skip connections to avoid the gradient vanishing. Finally, a convolution layer is applied to generate a residual image. 
The specific structure of our restoration network can be referred to in Figure \ref{fig:overview}. 

The restoration network takes the degraded image $I$ and degradation representation $r$ as input and performs adaptive processing to generate a high-quality image $R$.
In the training phase, to optimize the parameters in the network, we adopt two commonly used losses, smoothL1 \cite{girshick2015fast} and SSIM loss, which can measure the discrepancy of the restored result $R$ and ground truth image $Y$ at the pixel-wise level and patch-wise level, respectively. The optimization objective of our restoration network is the combination of two losses, which is formulated as follows:
\begin{equation}
    \mathcal{L}_{res} = \mathcal{L}_{smoothL_1}(R,Y) + \alpha \mathcal{L}_{ssim}(R,Y) ,
\end{equation}
where $\alpha$ is a hyper-parameter to balance the two losses. 
In the first stage, the DRN is required to be optimized by cross-entropy loss, therefore, the total loss in our method can be summarized as follows:
\begin{equation}
    \mathcal{L}_{total} = \mathcal{L}_{cls} + \mathcal{L}_{res} .
\end{equation}
In the second stage, the degradation representation $r$ is produced by detached DRN, and we focus on the optimization of the restoration network. Hence, the total loss in the second stage is $\mathcal{L}_{res}$.

\section{Experiments}
In this section, we conduct extensive experiments to verify the effectiveness of the proposed method. We first introduce the dataset and implementation details. Then we present the comparison with state-of-the-art methods across distortion types and across distortion levels. Finally, we conduct an ablation study to evaluate the effect of each component in our network and give a visualization of the hierarchical degradation representation.

\begin{figure*}
    \centering
  \includegraphics[width=\linewidth,trim=5 5 5 5,clip]{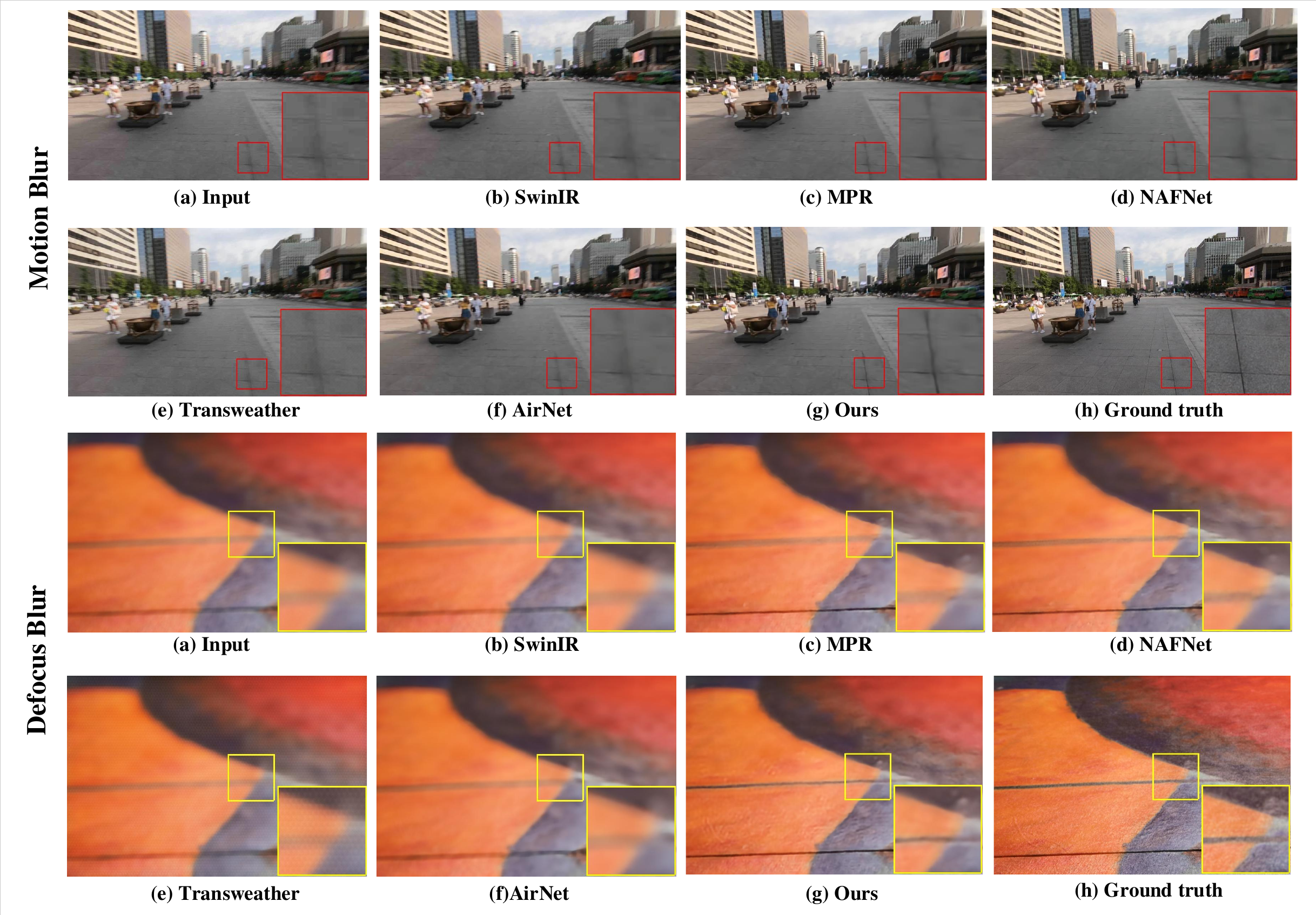}
  \caption{Visual comparison of our results and other state-of-the-art methods on blurry images from RED4\cite{Nah_2019_CVPR_Workshops_REDS} and DPDD\cite{abuolaim2020defocus} datasets. Our method achieves sharper details. Zoom in for better visualization.}
  \label{fig:visual_comparison_deblurring}
  \vspace{-0.3cm}
\end{figure*}

\subsection{Datasets and Implementation Details}

\textbf{Datasets.} As our work primarily focuses on multi-degradation restoration, the dataset used should contain adverse types of degraded images. For this purpose, we train our network on a combination of different degradation datasets, including RED4\cite{Nah_2019_CVPR_Workshops_REDS} for deblurring and jpeg-compression removal, SIDD\cite{abdelhamed2018high} for denoising, LOL \cite{Chen2018Retinex} for low-light enhancement, and DPDD \cite{abuolaim2020defocus} for defocus deblurring. The training data consists of 1920 images uniformly sampled from the four datasets. And we randomly sample 539 images for testing to validate the effectiveness of our method. Moreover, to measure the network's performance on images with different degradation levels, we follow \cite{li2022all} to train our network on WED \cite{ma2016waterloo} and test it on synthetic noisy images with different noise levels from CBSD68 \cite{martin2001database}.

\textbf{Implementation Details.} We adopt NAFNet \cite{chen2022simple} as our restoration backbone, as it has demonstrated remarkable performance and computational advantages across multiple restoration tasks. We implement our approach using Pytorch framework and train the network on two NVIDIA A100 GPUs in a distributed manner. An AdamW \cite{loshchilov2017fixing} optimizer is adopted to optimize network parameters. The learning rate is initialized to 5e-4 and decreased with the CosineAnnealingLR decay strategy. The network is trained for 600 epochs. The batch size is set to 28 and the patch size is 256. In the first stage, the clustering algorithm is conducted every 150 epochs.

\begin{figure*}[h]
    \centering
  \includegraphics[width=\linewidth,trim=5 5 5 5,clip]{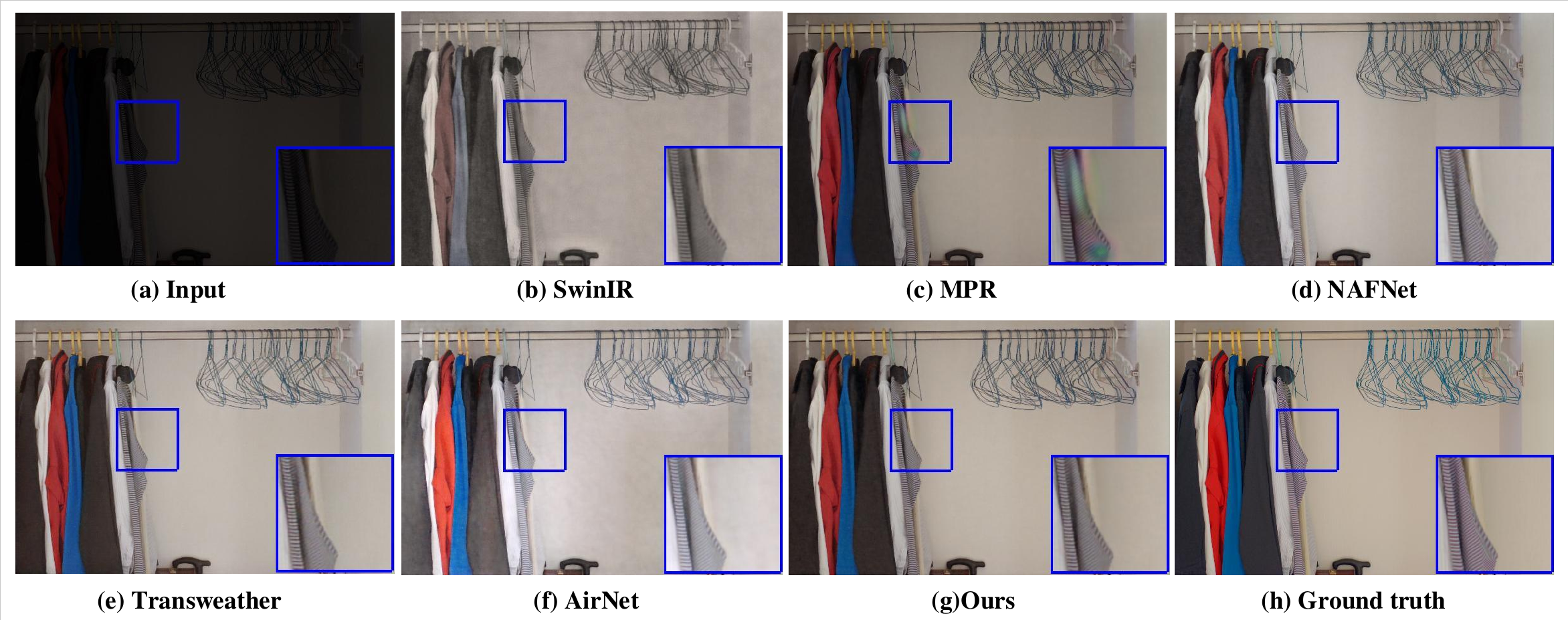}
  \caption{Visual comparison of our results and other state-of-the-art methods on low-light images from the LOL dataset \cite{Chen2018Retinex}. Our network effectively enhances the low-light image and generates harmonious tones. Zoom in for better visualization.}
  \label{fig:visual_comparison_enhancement}
  \vspace{-0.2cm}
\end{figure*}

\subsection{Comparisons cross Degradation Types}

\textbf{Compared Methods.} In this section, we evaluate our method with the state-of-the-art methods of image restoration. Initially, we selected some restoration approaches that are tailored for individual restoration tasks, including deblurring, denoising, low-light image enhancement, and defocus. Those methods like MPR \cite{zamir2021multi}, NAFNet \cite{chen2022simple}, which are CNN-based methods, and SwinIR \cite{liang2021swinir} is a transformer-based method. Besides, we also compare the performance of our method with other all-in-one approaches, like TransWeather \cite{valanarasu2022transweather} and AirNet \cite{li2022all}. For a fair competition with the compared methods, we follow them to conduct experiments with default optimal hyperparameters and settings. During the experiments, all the comparative methods were trained and tested on multi-degraded datasets in order to explore and compare their performance in handling multi-degraded images.

\textbf{Quantitative Comparison.} In this comparison, we adopt three commonly used in image restoration reference-based image quality assessment metrics, PSNR, SSIM, and LPIPS \cite{zhang2018unreasonable} to evaluate the quality of restored images. Better results are indicated by higher values of PSNR and SSIM, while LPIPS is the opposite. During the test, we calculate the metrics not only on all testing degraded images but also on each type of degradation according to the dataset type. As illustrated in Table \ref{tab: sota-results}, our method performs better than other methods in terms of the average metrics on all types of degraded data. Additionally, when classified by types of degradation, our method achieves the best or second-best results on different types of degraded images, suggesting that the performance improvement of our method can effectively distinguish and represent each type of degradation, rather than relying on the improvement on a single type of degradation. We also measured the computational costs of different models using a 3x256x256 image as input. It is evident that our method exhibits significantly reduced computational cost compared to AirNe t\cite{li2022all}, MPR \cite{zamir2021multi}, and SwinIR \cite{liang2021swinir}. While there is an increase in computational cost compared to NAFNet \cite{chen2022simple} and TransWeather \cite{valanarasu2022transweather}, our method holds an advantage in handling multiple degradations.

\begin{figure}
    \centering
  \includegraphics[width=\linewidth,trim=5 5 5 5,clip]{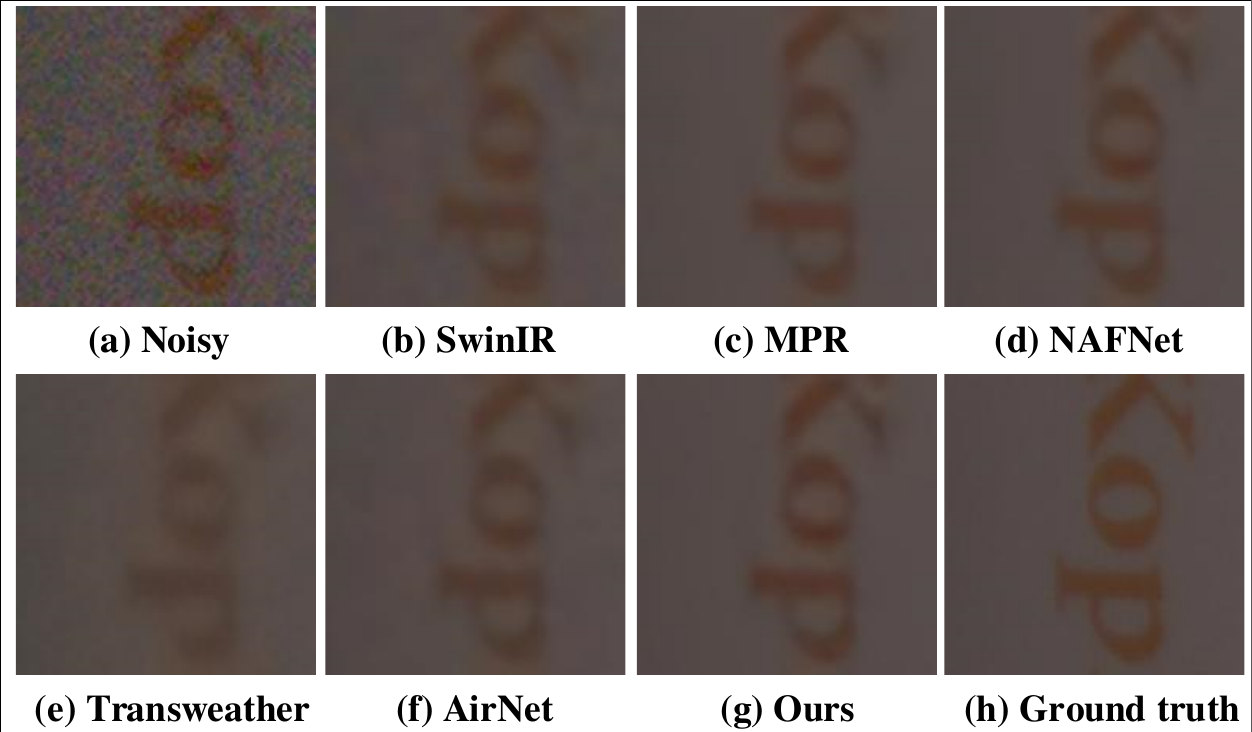}
  \caption{Visual comparison of our results and other state-of-the-art methods on a noisy image from SIDD \cite{abdelhamed2018high}. Our method produces results with less noise and sharper details. Zoom in for better visualization.}
  \label{fig:visual_comparison_denoising}
  \vspace{-0.2cm}
\end{figure}

\textbf{Visual Comparison.}
We conduct a visual comparison of our method and several state-of-the-art techniques on the aforementioned types of degraded images.
Figure\ref{fig:visual_comparison_deblurring} presents a visual comparison against other image restoration methods on RED4 \cite{Nah_2019_CVPR_Workshops_REDS} and DPDD \cite{abuolaim2020defocus} datasets, and it demonstrates that our method is capable of restoring the intricate details of the image while the image content is corrupted seriously by blur. Meanwhile, the results on LOL \cite{Chen2018Retinex} dataset are shown in Figure \ref{fig:visual_comparison_enhancement}, our method effectively removes artifacts from low-light images when enhancing them, ensuring that the processed images have good visual quality. Figure \ref{fig:visual_comparison_denoising} is the comparison of denoising results on SIDD \cite{abdelhamed2018high} dataset, our approach can remove the noise in distorted images and recover a high-quality image.
There are more results about the multi-degradation experiment, which can be found in the supplement material.

\begin{table}[htbp]\small
\caption{Quantitative comparison of our method and other state-of-the-art methods on denoising dataset CBSD68 \cite{martin2001database}. The best results are shown in boldface}
\setlength\tabcolsep{2pt}
\begin{tabular}{l|ll|ll|ll|ll}
\hline
\multirow{2}{*}{Model}          & \multicolumn{2}{c|}{Average}                  & \multicolumn{2}{c}{$\sigma$=15} &\multicolumn{2}{c}{$\sigma$=25} &\multicolumn{2}{c}{$\sigma$=50}                          \\ \cline{2-9} 
                                & \multicolumn{1}{c}{PSNR} & \multicolumn{1}{c|}{SSIM}  &
                                 \multicolumn{1}{c}{PSNR} & \multicolumn{1}{c|}{SSIM} &
                                 \multicolumn{1}{c}{PSNR} & \multicolumn{1}{c|}{SSIM}  &
                                 \multicolumn{1}{c}{PSNR} & \multicolumn{1}{c}{SSIM}  \\ \hline
    BM3D\cite{dabov2007image} & 30.54 &0.8505 &33.52 &0.9215& 30.71 &0.8672 & 27.38 & 0.7627  \\
    DnCNN\cite{zhang2017beyond}    & 31.03 & 0.8672 & 33.90 & 0.9290 & 31.24 & 0.8830 & 27.95 & 0.7896\\
    IRCNN \cite{zhang2017learning} & 30.98 & 0.8669 & 33.87&0.9285 &31.18 & 0.8824 & 27.88 & 0.7898 \\
    DL \cite{fan2019general} & 30.10& 0.8440 & 33.25 & 0.9225 & 30.38 & 0.8679 & 26.68 &0.7415 \\
    FFDNet\cite{zhang2018ffdnet}&31.01 & 0.8666 & 33.87& 0.9290& 31.21 & 0.8821& 27.96&0.7887 \\
    MPR\cite{zamir2021multi} & 31.15 & 0.8747 &34.01&0.9334 &31.34 &0.8892 &28.10 &0.8014\\
    
    Ours &\textbf{31.21} & \textbf{0.8783} & \textbf{34.05} & \textbf{0.9357} & \textbf{31.40} &\textbf{0.8929} &\textbf{28.17}& \textbf{0.8064}\\
\hline
\end{tabular}
\label{tab: comparison on CSBD68}
\vspace{-0.3cm}
\end{table}

\subsection{Comparison cross Degradation Levels}
To evaluate our network's performance on distorted images with different degradation levels, we compare the denoising results of our method and other state-of-the-art denoising methods, including BM3D \cite{dabov2007image}, DnCNN \cite{zhang2017beyond}, IRCNN \cite{zhang2017learning}, DL \cite{fan2019general}, FFDNet \cite{zhang2018ffdnet}, and MPR \cite{zamir2021multi}.
The noisy images are synthetic by adding White Gaussian noise with different levels (\textit{i.e.} $\sigma=15,25,50$) to clear images.
The comparison result is reported in Table \ref{tab: comparison on CSBD68}. Our approach exhibits superior  performance compared to other denoising methods and delivers optimal results in quantitative evaluations. This highlights the versatility of our method in restoring images degraded at varying levels.

\subsection{Ablation Study}

\begin{figure}[htbp]
\vspace{-0.4cm}
    \centering
  \includegraphics[width=0.8\linewidth]{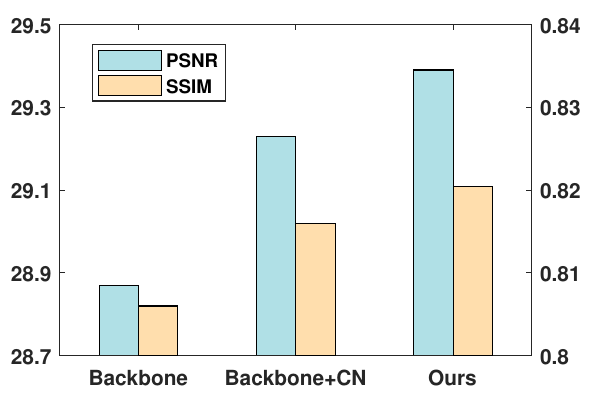}
  \caption{Performance comparison of models with/without the supervision of degradation information on multi-degraded images.}
  \label{fig:supervision_labels}
  \vspace{-0.3cm}
\end{figure}

\textbf{Impact of Supervision on Degradation Types.} To investigate the impact of degradation category supervision on the restoration network, we retrain some separate networks. As listed in Figure \ref{fig:supervision_labels}, the backbone model is a restoration network that does not utilize any degradation information. And backbone+CN is a restoration network that uses a classification network trained with degradation-type supervision and utilizes the classification results.  The comparative results in Figure \ref{fig:supervision_labels} show that the performance of networks that use degradation representation for processing multiple degradations has been significantly improved. However, even with the supervision of degradation types in the classification network, it cannot surpass our proposed method of using hierarchical degradation representation for restoration. 
This implies that degradation types are not mutually exclusive in multi-degraded image restoration, and exploring the relationships between degradations and leveraging them for restoration is highly necessary.

\textbf{Impact of Hierarchical Representation Layers.} We evaluate the performance of  networks with different layers in hierarchical degradation representation. The visual results are shown in Figure \ref{fig:visualization}.
From the results shown in Table \ref{tab:FTB}, it can be observed that as the number of layers in the tree structure increases, the restoration performance also improves with the enhancement of representation capacity. When using the two-stage strategy, the degradation representation provided by the fixed DRN allows the training to focus on RN. As a result, the performance of the restoration network is further improved.

\begin{table}[htbp]
    \caption{Ablation experiments for representation layers and FTB on multi-degraded datasets.}
    \begin{tabular}{c|cccc}
    \hline
     Number of layers & 1 layer & 2 layers  & 3 layers & 4 layers \\ \hline 
     PSNR &29.15 &29.23 & 29.26& 29.28 \\
     SSIM &0.8163 & 0.8172& 0.8179& 0.8185\\ \hline
     Models& w/o FTB & w/o DSLN & w/o GM & full \\ \hline
     PSNR & 28.87& 29.17& 29.22 & 29.39 \\
     SSIM &0.8060 & 0.8149&0.8166 & 0.8204\\ \hline

    \end{tabular}

    \label{tab:FTB}
\end{table}

\textbf{Impact of FTB.} We also conduct an ablation study to validate the effectiveness of FTB in our network. We remove FTB, DSLN, and GM separately in our model, to observe the effect of each component. The results are shown in Table \ref{tab:FTB}. Removing either DLSN or GM leads to a performance drop in our network.
However, as both have the ability to refine degraded image features by utilizing degradation information, the absence of one does not hinder the other from being effective. FTB's ability to modulate image features is maximized when both are used.

\subsection{Visualization}

\begin{figure}[htbp]
    \centering\includegraphics[width=0.99\linewidth,trim=1 1 1 1,clip]
    {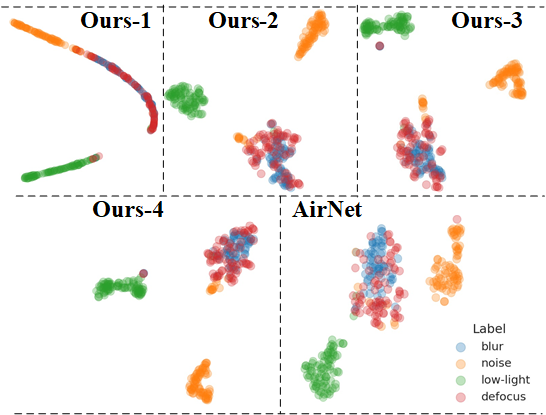}
  \caption{Visual comparison of degradation representation by t-SNE. The number in "Ours-number" represents the representation layers of AMIRNet.}
  \label{fig:visualization}
  \vspace{-0.3cm}
\end{figure}

We randomly sample degraded images from the training set and extract their degradation features through DRN. By utilizing t-SNE on the degradation representations, we can evaluate the performance of degradation representation in low-dimension embedding space. The visualized result is presented in Figure \ref{fig:visualization}. From the illustration, we can see that the hierarchical degradation representations generated by our method exhibit greater inter-class separation and intra-class compactness compared to the results of AirNet \cite{li2022all}. This result is consistent with our previous conjecture that tighter constraints on samples with similar degradation types can help improve the performance of the restoration network.

\section{Conclusion}

We present an all-in-one multi-degradation image restoration network, AMIRNet, an efficient and practical method to handle multi-degraded images. By progressively constructing a hierarchical degradation representation, AMIRNet can effectively model the similarity and differences among degradations. To overcome the domain gaps and  sufficiently utilize the degradation information, AMIRNet uses FTB to refine features with the guidance of the degradation representation. Extensive experimental results show that AMIRNet outperforms other state-of-the-art methods in multi-degradation restoration. The effectiveness of our approach is also verified by ablation studies and visualization.

\begin{acks}
This work was supported by National Science Foundation of China under Grant No.U19B2037 and No.61901384, Natural Science Basic Research Program of Shaanxi Province (Program No.2021JCW-03, No.2023-JC-QN-0685).
\end{acks}
\bibliographystyle{ACM-Reference-Format}
\balance
\bibliography{sample-base}


\begin{thebibliography}{43}


\ifx \showCODEN    \undefined \def \showCODEN     #1{\unskip}     \fi
\ifx \showDOI      \undefined \def \showDOI       #1{#1}\fi
\ifx \showISBNx    \undefined \def \showISBNx     #1{\unskip}     \fi
\ifx \showISBNxiii \undefined \def \showISBNxiii  #1{\unskip}     \fi
\ifx \showISSN     \undefined \def \showISSN      #1{\unskip}     \fi
\ifx \showLCCN     \undefined \def \showLCCN      #1{\unskip}     \fi
\ifx \shownote     \undefined \def \shownote      #1{#1}          \fi
\ifx \showarticletitle \undefined \def \showarticletitle #1{#1}   \fi
\ifx \showURL      \undefined \def \showURL       {\relax}        \fi
\providecommand\bibfield[2]{#2}
\providecommand\bibinfo[2]{#2}
\providecommand\natexlab[1]{#1}
\providecommand\showeprint[2][]{arXiv:#2}

\bibitem[Abdelhamed et~al\mbox{.}(2018)]%
        {abdelhamed2018high}
\bibfield{author}{\bibinfo{person}{Abdelrahman Abdelhamed},
  \bibinfo{person}{Stephen Lin}, {and} \bibinfo{person}{Michael~S Brown}.}
  \bibinfo{year}{2018}\natexlab{}.
\newblock \showarticletitle{A high-quality denoising dataset for smartphone
  cameras}. In \bibinfo{booktitle}{\emph{Proceedings of the IEEE conference on
  computer vision and pattern recognition}}. \bibinfo{pages}{1692--1700}.
\newblock


\bibitem[Abuolaim and Brown(2020)]%
        {abuolaim2020defocus}
\bibfield{author}{\bibinfo{person}{Abdullah Abuolaim} {and}
  \bibinfo{person}{Michael~S Brown}.} \bibinfo{year}{2020}\natexlab{}.
\newblock \showarticletitle{Defocus deblurring using dual-pixel data}. In
  \bibinfo{booktitle}{\emph{Computer Vision--ECCV 2020: 16th European
  Conference, Glasgow, UK, August 23--28, 2020, Proceedings, Part X 16}}.
  Springer, \bibinfo{pages}{111--126}.
\newblock


\bibitem[Chang et~al\mbox{.}(2019)]%
        {chang2019domain}
\bibfield{author}{\bibinfo{person}{Woong-Gi Chang}, \bibinfo{person}{Tackgeun
  You}, \bibinfo{person}{Seonguk Seo}, \bibinfo{person}{Suha Kwak}, {and}
  \bibinfo{person}{Bohyung Han}.} \bibinfo{year}{2019}\natexlab{}.
\newblock \showarticletitle{Domain-specific batch normalization for
  unsupervised domain adaptation}. In \bibinfo{booktitle}{\emph{Proceedings of
  the IEEE/CVF conference on Computer Vision and Pattern Recognition}}.
  \bibinfo{pages}{7354--7362}.
\newblock


\bibitem[Chen et~al\mbox{.}(2018)]%
        {chen2018learning}
\bibfield{author}{\bibinfo{person}{Chen Chen}, \bibinfo{person}{Qifeng Chen},
  \bibinfo{person}{Jia Xu}, {and} \bibinfo{person}{Vladlen Koltun}.}
  \bibinfo{year}{2018}\natexlab{}.
\newblock \showarticletitle{Learning to see in the dark}. In
  \bibinfo{booktitle}{\emph{Proceedings of the IEEE conference on computer
  vision and pattern recognition}}. \bibinfo{pages}{3291--3300}.
\newblock


\bibitem[Chen et~al\mbox{.}(2022a)]%
        {chen2022simple}
\bibfield{author}{\bibinfo{person}{Liangyu Chen}, \bibinfo{person}{Xiaojie
  Chu}, \bibinfo{person}{Xiangyu Zhang}, {and} \bibinfo{person}{Jian Sun}.}
  \bibinfo{year}{2022}\natexlab{a}.
\newblock \showarticletitle{Simple baselines for image restoration}. In
  \bibinfo{booktitle}{\emph{Computer Vision--ECCV 2022: 17th European
  Conference, Tel Aviv, Israel, October 23--27, 2022, Proceedings, Part VII}}.
  Springer, \bibinfo{pages}{17--33}.
\newblock


\bibitem[Chen et~al\mbox{.}(2021a)]%
        {chen2021hinet}
\bibfield{author}{\bibinfo{person}{Liangyu Chen}, \bibinfo{person}{Xin Lu},
  \bibinfo{person}{Jie Zhang}, \bibinfo{person}{Xiaojie Chu}, {and}
  \bibinfo{person}{Chengpeng Chen}.} \bibinfo{year}{2021}\natexlab{a}.
\newblock \showarticletitle{Hinet: Half instance normalization network for
  image restoration}. In \bibinfo{booktitle}{\emph{Proceedings of the IEEE/CVF
  Conference on Computer Vision and Pattern Recognition}}.
  \bibinfo{pages}{182--192}.
\newblock


\bibitem[Chen et~al\mbox{.}(2021b)]%
        {chen2021blind}
\bibfield{author}{\bibinfo{person}{Liang Chen}, \bibinfo{person}{Jiawei Zhang},
  \bibinfo{person}{Songnan Lin}, \bibinfo{person}{Faming Fang}, {and}
  \bibinfo{person}{Jimmy~S Ren}.} \bibinfo{year}{2021}\natexlab{b}.
\newblock \showarticletitle{Blind deblurring for saturated images}. In
  \bibinfo{booktitle}{\emph{Proceedings of the IEEE/CVF Conference on Computer
  Vision and Pattern Recognition}}. \bibinfo{pages}{6308--6316}.
\newblock


\bibitem[Chen et~al\mbox{.}(2022b)]%
        {chen2022learning}
\bibfield{author}{\bibinfo{person}{Wei-Ting Chen}, \bibinfo{person}{Zhi-Kai
  Huang}, \bibinfo{person}{Cheng-Che Tsai}, \bibinfo{person}{Hao-Hsiang Yang},
  \bibinfo{person}{Jian-Jiun Ding}, {and} \bibinfo{person}{Sy-Yen Kuo}.}
  \bibinfo{year}{2022}\natexlab{b}.
\newblock \showarticletitle{Learning multiple adverse weather removal via
  two-stage knowledge learning and multi-contrastive regularization: Toward a
  unified model}. In \bibinfo{booktitle}{\emph{Proceedings of the IEEE/CVF
  Conference on Computer Vision and Pattern Recognition}}.
  \bibinfo{pages}{17653--17662}.
\newblock


\bibitem[Dabov et~al\mbox{.}(2007)]%
        {dabov2007image}
\bibfield{author}{\bibinfo{person}{Kostadin Dabov}, \bibinfo{person}{Alessandro
  Foi}, \bibinfo{person}{Vladimir Katkovnik}, {and} \bibinfo{person}{Karen
  Egiazarian}.} \bibinfo{year}{2007}\natexlab{}.
\newblock \showarticletitle{Image denoising by sparse 3-D transform-domain
  collaborative filtering}.
\newblock \bibinfo{journal}{\emph{IEEE Transactions on image processing}}
  \bibinfo{volume}{16}, \bibinfo{number}{8} (\bibinfo{year}{2007}),
  \bibinfo{pages}{2080--2095}.
\newblock


\bibitem[Fan et~al\mbox{.}(2019)]%
        {fan2019general}
\bibfield{author}{\bibinfo{person}{Qingnan Fan}, \bibinfo{person}{Dongdong
  Chen}, \bibinfo{person}{Lu Yuan}, \bibinfo{person}{Gang Hua},
  \bibinfo{person}{Nenghai Yu}, {and} \bibinfo{person}{Baoquan Chen}.}
  \bibinfo{year}{2019}\natexlab{}.
\newblock \showarticletitle{A general decoupled learning framework for
  parameterized image operators}.
\newblock \bibinfo{journal}{\emph{IEEE transactions on pattern analysis and
  machine intelligence}} \bibinfo{volume}{43}, \bibinfo{number}{1}
  (\bibinfo{year}{2019}), \bibinfo{pages}{33--47}.
\newblock


\bibitem[Girshick(2015)]%
        {girshick2015fast}
\bibfield{author}{\bibinfo{person}{Ross Girshick}.}
  \bibinfo{year}{2015}\natexlab{}.
\newblock \showarticletitle{Fast r-cnn}. In
  \bibinfo{booktitle}{\emph{Proceedings of the IEEE international conference on
  computer vision}}. \bibinfo{pages}{1440--1448}.
\newblock


\bibitem[Gong et~al\mbox{.}(2017)]%
        {gong2017motion}
\bibfield{author}{\bibinfo{person}{Dong Gong}, \bibinfo{person}{Jie Yang},
  \bibinfo{person}{Lingqiao Liu}, \bibinfo{person}{Yanning Zhang},
  \bibinfo{person}{Ian Reid}, \bibinfo{person}{Chunhua Shen},
  \bibinfo{person}{Anton Van Den~Hengel}, {and} \bibinfo{person}{Qinfeng Shi}.}
  \bibinfo{year}{2017}\natexlab{}.
\newblock \showarticletitle{From motion blur to motion flow: A deep learning
  solution for removing heterogeneous motion blur}. In
  \bibinfo{booktitle}{\emph{Proceedings of the IEEE conference on computer
  vision and pattern recognition}}. \bibinfo{pages}{2319--2328}.
\newblock


\bibitem[Gu et~al\mbox{.}(2014)]%
        {gu2014weighted}
\bibfield{author}{\bibinfo{person}{Shuhang Gu}, \bibinfo{person}{Lei Zhang},
  \bibinfo{person}{Wangmeng Zuo}, {and} \bibinfo{person}{Xiangchu Feng}.}
  \bibinfo{year}{2014}\natexlab{}.
\newblock \showarticletitle{Weighted nuclear norm minimization with application
  to image denoising}. In \bibinfo{booktitle}{\emph{Proceedings of the IEEE
  conference on computer vision and pattern recognition}}.
  \bibinfo{pages}{2862--2869}.
\newblock


\bibitem[Guo et~al\mbox{.}(2019)]%
        {guo2019toward}
\bibfield{author}{\bibinfo{person}{Shi Guo}, \bibinfo{person}{Zifei Yan},
  \bibinfo{person}{Kai Zhang}, \bibinfo{person}{Wangmeng Zuo}, {and}
  \bibinfo{person}{Lei Zhang}.} \bibinfo{year}{2019}\natexlab{}.
\newblock \showarticletitle{Toward convolutional blind denoising of real
  photographs}. In \bibinfo{booktitle}{\emph{Proceedings of the IEEE/CVF
  conference on computer vision and pattern recognition}}.
  \bibinfo{pages}{1712--1722}.
\newblock


\bibitem[Hu et~al\mbox{.}(2014)]%
        {hu2014deblurring}
\bibfield{author}{\bibinfo{person}{Zhe Hu}, \bibinfo{person}{Sunghyun Cho},
  \bibinfo{person}{Jue Wang}, {and} \bibinfo{person}{Ming-Hsuan Yang}.}
  \bibinfo{year}{2014}\natexlab{}.
\newblock \showarticletitle{Deblurring low-light images with light streaks}. In
  \bibinfo{booktitle}{\emph{Proceedings of the IEEE Conference on Computer
  Vision and Pattern Recognition}}. \bibinfo{pages}{3382--3389}.
\newblock


\bibitem[Li et~al\mbox{.}(2022b)]%
        {li2022all}
\bibfield{author}{\bibinfo{person}{Boyun Li}, \bibinfo{person}{Xiao Liu},
  \bibinfo{person}{Peng Hu}, \bibinfo{person}{Zhongqin Wu},
  \bibinfo{person}{Jiancheng Lv}, {and} \bibinfo{person}{Xi Peng}.}
  \bibinfo{year}{2022}\natexlab{b}.
\newblock \showarticletitle{All-in-one image restoration for unknown
  corruption}. In \bibinfo{booktitle}{\emph{Proceedings of the IEEE/CVF
  Conference on Computer Vision and Pattern Recognition}}.
  \bibinfo{pages}{17452--17462}.
\newblock


\bibitem[Li et~al\mbox{.}(2022c)]%
        {li2022learning}
\bibfield{author}{\bibinfo{person}{Dasong Li}, \bibinfo{person}{Yi Zhang},
  \bibinfo{person}{Ka~Chun Cheung}, \bibinfo{person}{Xiaogang Wang},
  \bibinfo{person}{Hongwei Qin}, {and} \bibinfo{person}{Hongsheng Li}.}
  \bibinfo{year}{2022}\natexlab{c}.
\newblock \showarticletitle{Learning Degradation Representations for Image
  Deblurring}. In \bibinfo{booktitle}{\emph{Computer Vision--ECCV 2022: 17th
  European Conference, Tel Aviv, Israel, October 23--27, 2022, Proceedings,
  Part XVIII}}. Springer, \bibinfo{pages}{736--753}.
\newblock


\bibitem[Li et~al\mbox{.}(2022a)]%
        {li2022learningsuper}
\bibfield{author}{\bibinfo{person}{Fengjun Li}, \bibinfo{person}{Xin Feng},
  \bibinfo{person}{Fanglin Chen}, \bibinfo{person}{Guangming Lu}, {and}
  \bibinfo{person}{Wenjie Pei}.} \bibinfo{year}{2022}\natexlab{a}.
\newblock \showarticletitle{Learning Generalizable Latent Representations for
  Novel Degradations in Super-Resolution}. In
  \bibinfo{booktitle}{\emph{Proceedings of the 30th ACM International
  Conference on Multimedia}}. \bibinfo{pages}{1797--1807}.
\newblock


\bibitem[Li et~al\mbox{.}(2020)]%
        {li2020all}
\bibfield{author}{\bibinfo{person}{Ruoteng Li}, \bibinfo{person}{Robby~T Tan},
  {and} \bibinfo{person}{Loong-Fah Cheong}.} \bibinfo{year}{2020}\natexlab{}.
\newblock \showarticletitle{All in one bad weather removal using architectural
  search}. In \bibinfo{booktitle}{\emph{Proceedings of the IEEE/CVF conference
  on computer vision and pattern recognition}}. \bibinfo{pages}{3175--3185}.
\newblock


\bibitem[Liang et~al\mbox{.}(2021)]%
        {liang2021swinir}
\bibfield{author}{\bibinfo{person}{Jingyun Liang}, \bibinfo{person}{Jiezhang
  Cao}, \bibinfo{person}{Guolei Sun}, \bibinfo{person}{Kai Zhang},
  \bibinfo{person}{Luc Van~Gool}, {and} \bibinfo{person}{Radu Timofte}.}
  \bibinfo{year}{2021}\natexlab{}.
\newblock \showarticletitle{Swinir: Image restoration using swin transformer}.
  In \bibinfo{booktitle}{\emph{Proceedings of the IEEE/CVF international
  conference on computer vision}}. \bibinfo{pages}{1833--1844}.
\newblock


\bibitem[Liu et~al\mbox{.}(2013)]%
        {liu2013single}
\bibfield{author}{\bibinfo{person}{Xinhao Liu}, \bibinfo{person}{Masayuki
  Tanaka}, {and} \bibinfo{person}{Masatoshi Okutomi}.}
  \bibinfo{year}{2013}\natexlab{}.
\newblock \showarticletitle{Single-image noise level estimation for blind
  denoising}.
\newblock \bibinfo{journal}{\emph{IEEE transactions on image processing}}
  \bibinfo{volume}{22}, \bibinfo{number}{12} (\bibinfo{year}{2013}),
  \bibinfo{pages}{5226--5237}.
\newblock


\bibitem[Loshchilov and Hutter(2017)]%
        {loshchilov2017fixing}
\bibfield{author}{\bibinfo{person}{Ilya Loshchilov} {and}
  \bibinfo{person}{Frank Hutter}.} \bibinfo{year}{2017}\natexlab{}.
\newblock \showarticletitle{Fixing weight decay regularization in adam}.
\newblock  (\bibinfo{year}{2017}).
\newblock


\bibitem[Luo et~al\mbox{.}(2015)]%
        {luo2015removing}
\bibfield{author}{\bibinfo{person}{Yu Luo}, \bibinfo{person}{Yong Xu}, {and}
  \bibinfo{person}{Hui Ji}.} \bibinfo{year}{2015}\natexlab{}.
\newblock \showarticletitle{Removing rain from a single image via
  discriminative sparse coding}. In \bibinfo{booktitle}{\emph{Proceedings of
  the IEEE international conference on computer vision}}.
  \bibinfo{pages}{3397--3405}.
\newblock


\bibitem[Ma et~al\mbox{.}(2016)]%
        {ma2016waterloo}
\bibfield{author}{\bibinfo{person}{Kede Ma}, \bibinfo{person}{Zhengfang
  Duanmu}, \bibinfo{person}{Qingbo Wu}, \bibinfo{person}{Zhou Wang},
  \bibinfo{person}{Hongwei Yong}, \bibinfo{person}{Hongliang Li}, {and}
  \bibinfo{person}{Lei Zhang}.} \bibinfo{year}{2016}\natexlab{}.
\newblock \showarticletitle{Waterloo exploration database: New challenges for
  image quality assessment models}.
\newblock \bibinfo{journal}{\emph{IEEE Transactions on Image Processing}}
  \bibinfo{volume}{26}, \bibinfo{number}{2} (\bibinfo{year}{2016}),
  \bibinfo{pages}{1004--1016}.
\newblock


\bibitem[Mairal et~al\mbox{.}(2007)]%
        {mairal2007sparse}
\bibfield{author}{\bibinfo{person}{Julien Mairal}, \bibinfo{person}{Michael
  Elad}, {and} \bibinfo{person}{Guillermo Sapiro}.}
  \bibinfo{year}{2007}\natexlab{}.
\newblock \showarticletitle{Sparse representation for color image restoration}.
\newblock \bibinfo{journal}{\emph{IEEE Transactions on image processing}}
  \bibinfo{volume}{17}, \bibinfo{number}{1} (\bibinfo{year}{2007}),
  \bibinfo{pages}{53--69}.
\newblock


\bibitem[Martin et~al\mbox{.}(2001)]%
        {martin2001database}
\bibfield{author}{\bibinfo{person}{David Martin}, \bibinfo{person}{Charless
  Fowlkes}, \bibinfo{person}{Doron Tal}, {and} \bibinfo{person}{Jitendra
  Malik}.} \bibinfo{year}{2001}\natexlab{}.
\newblock \showarticletitle{A database of human segmented natural images and
  its application to evaluating segmentation algorithms and measuring
  ecological statistics}. In \bibinfo{booktitle}{\emph{Proceedings Eighth IEEE
  International Conference on Computer Vision. ICCV 2001}},
  Vol.~\bibinfo{volume}{2}. IEEE, \bibinfo{pages}{416--423}.
\newblock


\bibitem[Nah et~al\mbox{.}(2019)]%
        {Nah_2019_CVPR_Workshops_REDS}
\bibfield{author}{\bibinfo{person}{Seungjun Nah}, \bibinfo{person}{Sungyong
  Baik}, \bibinfo{person}{Seokil Hong}, \bibinfo{person}{Gyeongsik Moon},
  \bibinfo{person}{Sanghyun Son}, \bibinfo{person}{Radu Timofte}, {and}
  \bibinfo{person}{Kyoung~Mu Lee}.} \bibinfo{year}{2019}\natexlab{}.
\newblock \showarticletitle{NTIRE 2019 Challenge on Video Deblurring and
  Super-Resolution: Dataset and Study}. In \bibinfo{booktitle}{\emph{CVPR
  Workshops}}.
\newblock


\bibitem[Pan et~al\mbox{.}(2016)]%
        {pan2016robust}
\bibfield{author}{\bibinfo{person}{Jinshan Pan}, \bibinfo{person}{Zhouchen
  Lin}, \bibinfo{person}{Zhixun Su}, {and} \bibinfo{person}{Ming-Hsuan Yang}.}
  \bibinfo{year}{2016}\natexlab{}.
\newblock \showarticletitle{Robust kernel estimation with outliers handling for
  image deblurring}. In \bibinfo{booktitle}{\emph{Proceedings of the IEEE
  Conference on Computer Vision and Pattern Recognition}}.
  \bibinfo{pages}{2800--2808}.
\newblock


\bibitem[Valanarasu et~al\mbox{.}(2022)]%
        {valanarasu2022transweather}
\bibfield{author}{\bibinfo{person}{Jeya Maria~Jose Valanarasu},
  \bibinfo{person}{Rajeev Yasarla}, {and} \bibinfo{person}{Vishal~M Patel}.}
  \bibinfo{year}{2022}\natexlab{}.
\newblock \showarticletitle{Transweather: Transformer-based restoration of
  images degraded by adverse weather conditions}. In
  \bibinfo{booktitle}{\emph{Proceedings of the IEEE/CVF Conference on Computer
  Vision and Pattern Recognition}}. \bibinfo{pages}{2353--2363}.
\newblock


\bibitem[Wang et~al\mbox{.}(2021)]%
        {wang2021unsupervised}
\bibfield{author}{\bibinfo{person}{Longguang Wang}, \bibinfo{person}{Yingqian
  Wang}, \bibinfo{person}{Xiaoyu Dong}, \bibinfo{person}{Qingyu Xu},
  \bibinfo{person}{Jungang Yang}, \bibinfo{person}{Wei An}, {and}
  \bibinfo{person}{Yulan Guo}.} \bibinfo{year}{2021}\natexlab{}.
\newblock \showarticletitle{Unsupervised degradation representation learning
  for blind super-resolution}. In \bibinfo{booktitle}{\emph{Proceedings of the
  IEEE/CVF Conference on Computer Vision and Pattern Recognition}}.
  \bibinfo{pages}{10581--10590}.
\newblock


\bibitem[Wei et~al\mbox{.}(2018)]%
        {Chen2018Retinex}
\bibfield{author}{\bibinfo{person}{Chen Wei}, \bibinfo{person}{Wenjing Wang},
  \bibinfo{person}{Wenhan Yang}, {and} \bibinfo{person}{Jiaying Liu}.}
  \bibinfo{year}{2018}\natexlab{}.
\newblock \showarticletitle{Deep Retinex Decomposition for Low-Light
  Enhancement}. In \bibinfo{booktitle}{\emph{British Machine Vision
  Conference}}.
\newblock


\bibitem[Xu et~al\mbox{.}(2017)]%
        {xu2017multi}
\bibfield{author}{\bibinfo{person}{Jun Xu}, \bibinfo{person}{Lei Zhang},
  \bibinfo{person}{David Zhang}, {and} \bibinfo{person}{Xiangchu Feng}.}
  \bibinfo{year}{2017}\natexlab{}.
\newblock \showarticletitle{Multi-channel weighted nuclear norm minimization
  for real color image denoising}. In \bibinfo{booktitle}{\emph{Proceedings of
  the IEEE international conference on computer vision}}.
  \bibinfo{pages}{1096--1104}.
\newblock


\bibitem[Xu et~al\mbox{.}(2013)]%
        {xu2013unnatural}
\bibfield{author}{\bibinfo{person}{Li Xu}, \bibinfo{person}{Shicheng Zheng},
  {and} \bibinfo{person}{Jiaya Jia}.} \bibinfo{year}{2013}\natexlab{}.
\newblock \showarticletitle{Unnatural l0 sparse representation for natural
  image deblurring}. In \bibinfo{booktitle}{\emph{Proceedings of the IEEE
  conference on computer vision and pattern recognition}}.
  \bibinfo{pages}{1107--1114}.
\newblock


\bibitem[Yan et~al\mbox{.}(2019)]%
        {yan2019attention}
\bibfield{author}{\bibinfo{person}{Qingsen Yan}, \bibinfo{person}{Dong Gong},
  \bibinfo{person}{Qinfeng Shi}, \bibinfo{person}{Anton van~den Hengel},
  \bibinfo{person}{Chunhua Shen}, \bibinfo{person}{Ian Reid}, {and}
  \bibinfo{person}{Yanning Zhang}.} \bibinfo{year}{2019}\natexlab{}.
\newblock \showarticletitle{Attention-guided network for ghost-free high
  dynamic range imaging}. In \bibinfo{booktitle}{\emph{Proceedings of the
  IEEE/CVF Conference on Computer Vision and Pattern Recognition}}.
  \bibinfo{pages}{1751--1760}.
\newblock


\bibitem[Yan et~al\mbox{.}(2023)]%
        {yan2023sharpformer}
\bibfield{author}{\bibinfo{person}{Qingsen Yan}, \bibinfo{person}{Dong Gong},
  \bibinfo{person}{Pei Wang}, \bibinfo{person}{Zhen Zhang},
  \bibinfo{person}{Yanning Zhang}, {and} \bibinfo{person}{Javen~Qinfeng Shi}.}
  \bibinfo{year}{2023}\natexlab{}.
\newblock \showarticletitle{SharpFormer: Learning Local Feature Preserving
  Global Representations for Image Deblurring}.
\newblock \bibinfo{journal}{\emph{IEEE Transactions on Image Processing}}
  (\bibinfo{year}{2023}).
\newblock


\bibitem[Yan et~al\mbox{.}(2017)]%
        {yan2017image}
\bibfield{author}{\bibinfo{person}{Yanyang Yan}, \bibinfo{person}{Wenqi Ren},
  \bibinfo{person}{Yuanfang Guo}, \bibinfo{person}{Rui Wang}, {and}
  \bibinfo{person}{Xiaochun Cao}.} \bibinfo{year}{2017}\natexlab{}.
\newblock \showarticletitle{Image deblurring via extreme channels prior}. In
  \bibinfo{booktitle}{\emph{Proceedings of the IEEE Conference on Computer
  Vision and Pattern Recognition}}. \bibinfo{pages}{4003--4011}.
\newblock


\bibitem[Zamir et~al\mbox{.}(2021)]%
        {zamir2021multi}
\bibfield{author}{\bibinfo{person}{Syed~Waqas Zamir}, \bibinfo{person}{Aditya
  Arora}, \bibinfo{person}{Salman Khan}, \bibinfo{person}{Munawar Hayat},
  \bibinfo{person}{Fahad~Shahbaz Khan}, \bibinfo{person}{Ming-Hsuan Yang},
  {and} \bibinfo{person}{Ling Shao}.} \bibinfo{year}{2021}\natexlab{}.
\newblock \showarticletitle{Multi-stage progressive image restoration}. In
  \bibinfo{booktitle}{\emph{Proceedings of the IEEE/CVF conference on computer
  vision and pattern recognition}}. \bibinfo{pages}{14821--14831}.
\newblock


\bibitem[Zhang et~al\mbox{.}(2022)]%
        {zhang2022exploring}
\bibfield{author}{\bibinfo{person}{Cheng Zhang}, \bibinfo{person}{Shaolin Su},
  \bibinfo{person}{Yu Zhu}, \bibinfo{person}{Qingsen Yan},
  \bibinfo{person}{Jinqiu Sun}, {and} \bibinfo{person}{Yanning Zhang}.}
  \bibinfo{year}{2022}\natexlab{}.
\newblock \showarticletitle{Exploring and evaluating image restoration
  potential in dynamic scenes}. In \bibinfo{booktitle}{\emph{Proceedings of the
  IEEE/CVF Conference on Computer Vision and Pattern Recognition}}.
  \bibinfo{pages}{2067--2076}.
\newblock


\bibitem[Zhang et~al\mbox{.}(2020)]%
        {zhang2020attention}
\bibfield{author}{\bibinfo{person}{Cheng Zhang}, \bibinfo{person}{Qingsen Yan},
  \bibinfo{person}{Yu Zhu}, \bibinfo{person}{Xianjun Li},
  \bibinfo{person}{Jinqiu Sun}, {and} \bibinfo{person}{Yanning Zhang}.}
  \bibinfo{year}{2020}\natexlab{}.
\newblock \showarticletitle{Attention-based network for low-light image
  enhancement}. In \bibinfo{booktitle}{\emph{2020 IEEE international conference
  on multimedia and expo (ICME)}}. IEEE, \bibinfo{pages}{1--6}.
\newblock


\bibitem[Zhang et~al\mbox{.}(2017a)]%
        {zhang2017beyond}
\bibfield{author}{\bibinfo{person}{Kai Zhang}, \bibinfo{person}{Wangmeng Zuo},
  \bibinfo{person}{Yunjin Chen}, \bibinfo{person}{Deyu Meng}, {and}
  \bibinfo{person}{Lei Zhang}.} \bibinfo{year}{2017}\natexlab{a}.
\newblock \showarticletitle{Beyond a gaussian denoiser: Residual learning of
  deep cnn for image denoising}.
\newblock \bibinfo{journal}{\emph{IEEE transactions on image processing}}
  \bibinfo{volume}{26}, \bibinfo{number}{7} (\bibinfo{year}{2017}),
  \bibinfo{pages}{3142--3155}.
\newblock


\bibitem[Zhang et~al\mbox{.}(2017b)]%
        {zhang2017learning}
\bibfield{author}{\bibinfo{person}{Kai Zhang}, \bibinfo{person}{Wangmeng Zuo},
  \bibinfo{person}{Shuhang Gu}, {and} \bibinfo{person}{Lei Zhang}.}
  \bibinfo{year}{2017}\natexlab{b}.
\newblock \showarticletitle{Learning deep CNN denoiser prior for image
  restoration}. In \bibinfo{booktitle}{\emph{Proceedings of the IEEE conference
  on computer vision and pattern recognition}}. \bibinfo{pages}{3929--3938}.
\newblock


\bibitem[Zhang et~al\mbox{.}(2018b)]%
        {zhang2018ffdnet}
\bibfield{author}{\bibinfo{person}{Kai Zhang}, \bibinfo{person}{Wangmeng Zuo},
  {and} \bibinfo{person}{Lei Zhang}.} \bibinfo{year}{2018}\natexlab{b}.
\newblock \showarticletitle{FFDNet: Toward a fast and flexible solution for
  CNN-based image denoising}.
\newblock \bibinfo{journal}{\emph{IEEE Transactions on Image Processing}}
  \bibinfo{volume}{27}, \bibinfo{number}{9} (\bibinfo{year}{2018}),
  \bibinfo{pages}{4608--4622}.
\newblock


\bibitem[Zhang et~al\mbox{.}(2018a)]%
        {zhang2018unreasonable}
\bibfield{author}{\bibinfo{person}{Richard Zhang}, \bibinfo{person}{Phillip
  Isola}, \bibinfo{person}{Alexei~A Efros}, \bibinfo{person}{Eli Shechtman},
  {and} \bibinfo{person}{Oliver Wang}.} \bibinfo{year}{2018}\natexlab{a}.
\newblock \showarticletitle{The unreasonable effectiveness of deep features as
  a perceptual metric}. In \bibinfo{booktitle}{\emph{Proceedings of the IEEE
  conference on computer vision and pattern recognition}}.
  \bibinfo{pages}{586--595}.
\newblock


\end{thebibliography}


\end{document}